\documentclass{article}
\usepackage{algorithm}
\usepackage{algorithmic}
\usepackage{graphicx}    % 处理图片
\usepackage{subcaption}  % 支持子图
\usepackage{rotating}
\usepackage{multirow}
\usepackage[english]{babel}
\usepackage[
  a4paper,
  top=2cm,
  bottom=2cm,
  left=3cm,
  right=3cm,
  marginparwidth=1.75cm
]{geometry}

\usepackage{indentfirst} % 强制首段缩进
\usepackage{ragged2e} % 两端对齐
\usepackage{amsfonts} % 数学符号

\justifying % 两端对齐
\usepackage{amsmath} % 数学公式
\usepackage{hyperref} % 超链接（放置在较后位置避免冲突）
\usepackage{natbib} % 参考文献

\title{HIT-ROCKET: Hadamard-vector Inner-product Transformer for ROCKET}

\author{
  Wang Hao$^{*}$ \quad
  \href{mailto:25b905039@stu.hit.edu.cn}{25b905039@stu.hit.edu.cn} \\
  Kuang Zhang \quad \href{mailto:zhangkuang@hit.edu.cn}{zhangkuang@hit.edu.cn} \\
  Hou Chengyu$^{\dagger}$ \quad \href{mailto:houcy@hit.edu.cn}{houcy@hit.edu.cn} \\
  Yuan Zhonghao \quad \href{mailto:yuanatp@qq.com}{yuanatp@qq.com} \\
  Tan Chenxing \quad \href{mailto:3424611356@qq.com}{3424611356@qq.com} \\
  Fu Weifeng \quad \href{mailto:2298000456@qq.com}{2298000456@qq.com} \\
  Zhu Yangying\quad \href{2023112080@stu.hit.edu.cn}{2023112080@stu.hit.edu.cn} \\
  \small School of Electronics and Information Engineering, \\
  \small Harbin Institute of Technology, Harbin 150001, China
}

\begin{document}
\maketitle % 生成标题页，避免手动换行导致的空白
\thanks{$^{*}$ First author. $^{\dagger}$ Corresponding author: houcy@hit.edu.cn.}

  \begin{abstract}
Time series classification holds broad application value in fields such as communications, information countermeasures, finance, and medicine. However, state-of-the-art (SOTA) time series classification methods—including HIVE-COTE, Proximity Forest, and TS-CHIEF—exhibit high computational complexity, accompanied by long parameter tuning and training cycles. In contrast, lightweight solutions are offered by methods like ROCKET (Random Convolutional Kernel Transform), yet significant room for improvement remains in aspects such as kernel selection and computational complexity.To address these challenges, we propose a feature extraction method based on the Hadamard convolutional transform. Specifically, column vectors or row vectors of the Hadamard matrix are employed as convolution kernels, with extended convolution lengths of different sizes. This improvement not only maintains full compatibility with the entire design of existing methods (e.g., ROCKET) but also further enhances the algorithm's computational efficiency, robustness, and adaptability, leveraging the orthogonality of Hadamard convolution kernels.Extensive experimental validations of the proposed method are conducted on datasets across multiple domains, with a focus on the large-scale open-source University of California, Riverside (UCR) time series dataset. A large number of experimental results demonstrate that our method achieves SOTA performance on all tested datasets: its F1-score is improved by at least 5\% compared to the ROCKET method. Meanwhile, when hyperparameters are configured identically, its training time is 50\% lower than that of miniROCKET (the fastest variant in the ROCKET family), facilitating deployment on ultra-low-power embedded devices. All our code is available on GitHub.

  \end{abstract}

  \noindent


  \section{INTRODUCTION}
Time series classification has extensive applications and significance across various domains, such as modulation recognition \cite{1}, financial analysis \cite{2}, and medical diagnosis \cite{3}. Over the past decades, numerous time series classification algorithms have been proposed, generally categorized into traditional feature extraction combined with statistical learning methods \cite{4}, and end-to-end deep learning methods that have emerged in recent years \cite{5}. However, these methods often require substantial computational resources, making them difficult to meet the demands of edge computing scenarios with cost, computing power, and power consumption bottlenecks, or special fields such as micro-nano robotics and aerospace.

ROCKET-family methods \cite{6,7,8,9} provide a solution with high accuracy and lightweight properties. In particular, miniROCKET \cite{7} determines the random convolutions in the ROCKET method as 84 fixed "random" convolution kernels composed of -1 and 2 (or -2 and 1), achieving the highest computational efficiency. Nevertheless, we found that directly fixing the weights of random convolution kernels as combinations of -1 and 2 may introduce issues such as mean shift due to couplings between kernels. This can amplify noise in certain scenarios, thereby degrading classification performance. The proposed Hadamard convolution, which replaces the "arbitrarily specified" convolutions in other ROCKET-family methods, not only has more reliable mathematical support but also further improves computational efficiency—making it more suitable for deployment on ultra-low-power embedded devices such as microcontrollers and FPGAs.

Specifically, similar to miniROCKET, HIT-ROCKET also uses dilated convolution to extract features from time series. The key difference lies in adopting Hadamard vectors as convolution kernel weights: this not only significantly reduces computational complexity but also enhances classification accuracy and noise resistance due to the irrelevance between kernels. In the main version, we only retain PPV (Positive Predictive Value) as the key feature. The low correlation between features and the small number of features allow us to not only use traditional logistic regression but also achieve excellent performance with tree-based models and SVM (Support Vector Machine) as classifiers. Fig. 1(A) presents the accuracy ranking. Meanwhile, compared with other ROCKET (especially miniROCKET) or no ROCKET methods , HIT-ROCKET maintains the shortest training and inference time due to the reduced number of convolution kernels, as shown in Fig. 1(B).

\begin{figure}[htbp]
    \centering
    \includegraphics[width=1\linewidth]{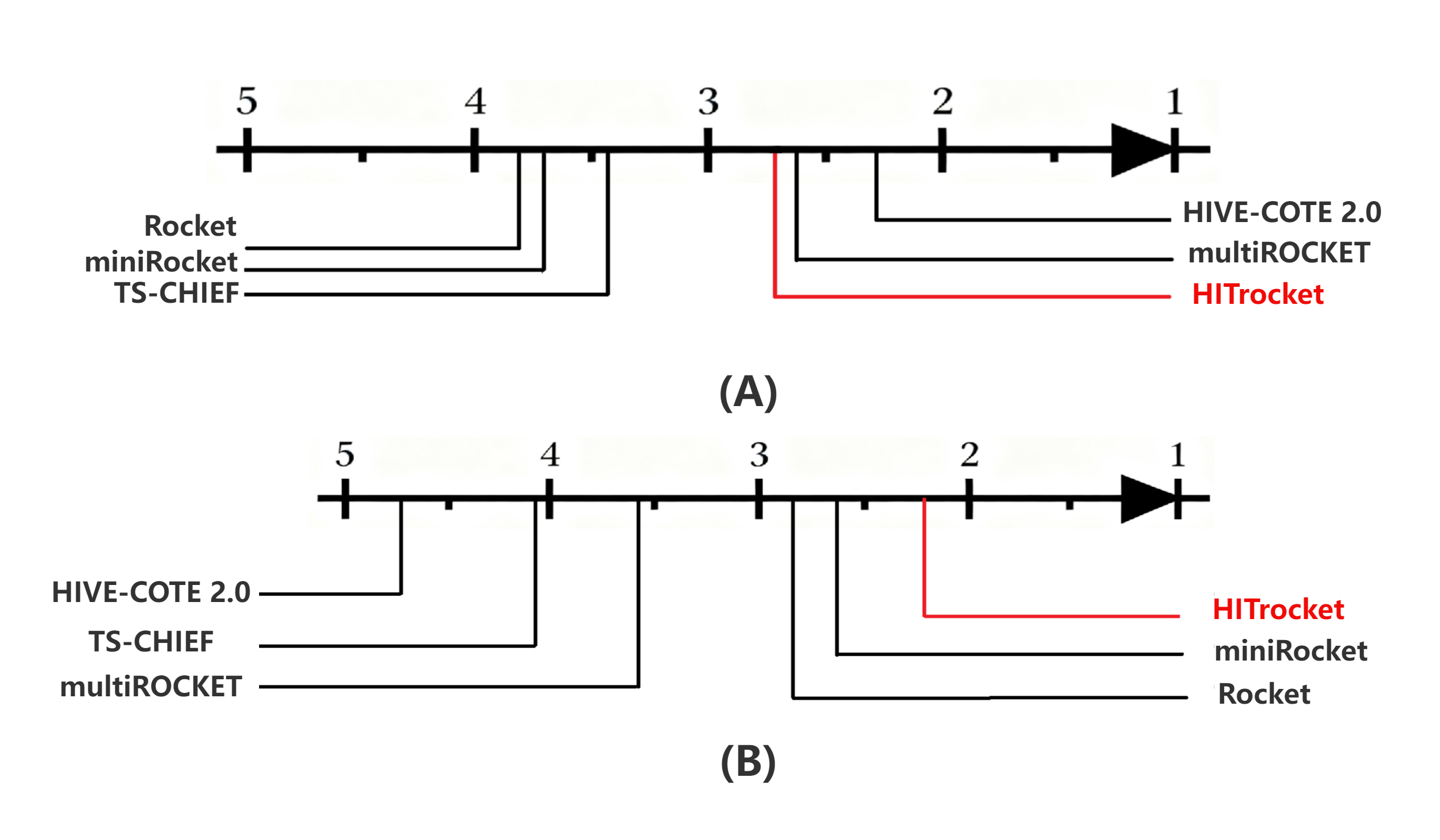}
    \caption{\small This figure presents the performance ranking results of experiments conducted on 20 datasets from the UCR archive, comparing the proposed method with 5 main competing methods (note: full coverage of HIVE-COTE experiments was not feasible due to computational constraints). (A) shows the accuracy ranking, where the proposed HIT-ROCKET method rankssthird; (B) displays the training and inference time ranking, with HIT-ROCKET ranking first.}
    \label{fig:1}
\end{figure}

The main contributions of this paper are as follows:
\begin{enumerate}
    \item We introduce a feature extraction method using Hadamard convolutional transform, and mathematically derive the principles behind performance improvements in aspects such as classification performance and noise resistance.
    \item Extensive experiments are conducted on the large-scale open-source University of California, Riverside (UCR) time series dataset, which covers multiple domains and diverse data characteristics. These experiments fully verify the universality and reliability of the proposed method.
    \item We provide additional extended extended feature extraction schemes, such as symplectic geometric reconstruction, non-uniform dilated convolution, and random dilated convolution. In some scenarios, these schemes can indeed further further further improve classification performance but may increase computational resource requirements and hardware implementation complexity.
    \item We provide an open-source implementation of the algorithm, which is compatible with the well-known scikit-learn machine learning library and supports PyTorch-CUDA for accelerating the transformation of large-scale datasets. Specifically, test demos deployed on ESP32 and RV1103 chips are also provided. The code will be made publicly available on GitHub upon acceptance, enabling researchers to easily apply our model to their own datasets and reproduce our experiments.
\end{enumerate}

The remaining structure of this paper is organized as follows: Section 2 reviews related research works. Section 3 introduces the Hadamard matrix and the mathematical formulation of the HIT-ROCKET algorithm, analyzes the upper bound of the generalization error of HIT-ROCKET from a mathematical perspective, and discusses the performance improvements brought by independent feature extraction. Section 4 elaborates on the specific implementation method of HIT-ROCKET. Section 5 presents extensive experiments on the UCR dataset to validate the effectiveness of the proposed method. Section 6 discusses the method and process of deploying HIT-ROCKET on ultra-low-power microcontrollers, and provides an outlook on future work. Section 7 concludes the paper.

\section{MATHEMATICAL PRINCIPLES}
\section{RELATED WORK}

Time series classification algorithms are mainly divided into traditional feature extraction combined with statistical learning methods, and deep learning methods that have emerged in recent years. The architectures adopted by deep learning methods mainly include convolutional neural networks, recurrent neural networks, attention-based transformers, graph neural networks, and their combinations. These methods have achieved encouraging success in some tasks, but they suffer from limitations such as requiring large amounts of samples, being sensitive to noise, demanding massive computational resources, and having poor interpretability. These drawbacks restrict the application of deep learning methods in edge scenarios with computational and power consumption bottlenecks. Therefore, non-deep learning methods have also received widespread attention and application. These non-deep learning methods generally extract features using multiple approaches and then perform classification based on these features using traditional statistical learning or Bayesian learning methods. They can be mainly categorized as follows \cite{4,10}:

\textbf{Full series}: These methods use the entire time series as features. The core idea is to define similarity distances between time series (e.g., Euclidean distance, dynamic time warping and its variants, time warping edit distance, etc.), and then use nearest neighbor algorithms (e.g., KNN) for classification \cite{11}.

\textbf{Series intervals}: Discriminative features are located in specific intervals of the time series and are computed using aggregation functions (e.g., mean, median, kurtosis, skewness, maximum value, etc.). Supervised classifiers are then trained on these computed features. To our knowledge, Time Series Forest (TSF) \cite{12} is the most popular algorithm in this category.

\textbf{Subsequence shapelet}: A shapelet is a "small shape" of a time series, i.e., a subsequence. This subsequence is a distinctive sub-segment in the time series data that can express the most significant characteristics of the temporal data. Instances are classified based on the shapelets they contain. This category includes methods such as Shapelet Decision Tree \cite{13}, ELIS++ \cite{14}, and SAST \cite{15}.

\textbf{Dictionary}: These methods are used when categories have the same discriminative patterns but with different frequencies per category. Typically, time series are converted into symbolic sequences, and then patterns (also called words) are extracted. The Bag of Symbolic-Fourier Approximation Symbols (BOSS) \cite{16} and its variant Contract BOSS \cite{17} are popular algorithms in this category.

\textbf{Transform domain}: These methods operate in the frequency domain, fractional Fourier domain, or wavelet domain, and can extract features that are difficult to detect in the time domain. Some popular techniques used here include power spectrum and autocorrelation function (ACF) \cite{18,19}.

\textbf{Hybrid}: These methods simultaneously process different types of features to leverage the advantages of each. The most well-known model in this category is HIVE-COTE \cite{20}, which consists of multiple components, each focusing on one type of feature (full series, intervals, shapelets, dictionary, or spectrum), and is an ensemble of classifiers. TS-CHIEF \cite{21} is a tree-based hybrid method that uses all the aforementioned features except shapelets. ROCKET \cite{6} is also considered a hybrid feature extraction model, which uses randomly generated convolution kernels to extract different types of features. Literature \cite{9} suggests that the ROCKET method is equivalent to dictionary methods to some extent.

Our method can be regarded as an improvement and supplement to ROCKET-family methods, especially miniROCKET. Since its proposal in 2020 \cite{6}, the ROCKET method has attracted widespread attention due to its ultra-short training time and excellent performance. The ROCKET family mainly includes three types: ROCKET, miniROCKET, and MultiROCKET. ROCKET is the earliest version, which samples weights from a Gaussian distribution and biases from a uniform distribution. It uses the proportion of positive values for pooling the convolution outputs to extract features from the input time series, and achieves good results using only linear classifiers such as ridge regression and logistic regression. As mentioned above, miniROCKET \cite{7} fixes both the convolution kernels and biases: its kernels have a fixed length of 9, composed of combinations of -2 and 1 (or 2 and -1), and biases are sampled by quantiles from the convolution outputs of several randomly selected sequence samples, further improving computational efficiency. MultiROCKET \cite{8}, building on miniROCKET, proposes using the difference sequence of the original sequence to open another transformation channel and introduces new features such as the position of the maximum positive value.

\section{PRELIMINARIES}

In this section, we first introduce the properties of Hadamard matrices and complex-domain Hadamard matrices. Then, we present the general formulation of the main HIT-ROCKET algorithm. Finally, we derive the matrix form of the proposed algorithm and its generalization error bound, thereby highlighting feature independence and the resulting performance improvements.

\subsection{Hadamard Matrix}

A Hadamard matrix is an important square matrix in mathematics and engineering, whose elements are either +1 or -1, and the correlation between any two rows (or columns) is zero. That is, the dot product of any two rows (or columns) is zero. In the proposed algorithm, we extract a column from the Hadamard matrix $ H_{N \times N} $ as a convolution kernel, referred to as a Hadamard vector $ h_i $.

\[
H_{N \times N}^T H_{N \times N} = N I
\]

where $ I $ is an $ N \times N $ identity matrix.

Hadamard matrices have the following important properties:

1. **Orthogonality**: The dot product of any two rows (or columns) is zero, making Hadamard matrices useful for orthogonal coding and multiplexing in signal processing.
2. **Symmetry**: The transpose of a Hadamard matrix equals its inverse, i.e., $ H^T = H^{-1} $, which is highly useful in matrix operations.
3. **Maximum determinant**: Among all square matrices with elements ±1, the absolute value of the determinant of a Hadamard matrix reaches the maximum.
4. **Existence condition**: A Hadamard matrix exists if and only if its order is 1, 2, or a multiple of 4. This condition limits its application scope but provides a basis for its construction.

In the complex domain, the definition of Hadamard matrices can be extended to matrices with complex elements, where the inner product of any two rows (or columns) is zero. The inner product of complex numbers $ z_1 = a_1 + {\rm j}b_1 $ and $ z_2 = a_2 + {\rm j}b_2 $ can be expressed as $ \text{Real}\{z_1^* z_2\} = a_1a_2 + b_1b_2 $. Thus,

\[
\text{Real}\{H_{N \times N}^T H_{N \times N}\} = N I
\]

Two example Hadamard matrices are given in Fig. 2.

\subsection{Preliminary Formulation of HIT-ROCKET}

Each column of an $ N $-order Hadamard matrix is extracted as the weight of a convolution kernel. The convolution (correlation) output $ X $ with a time series $ S $ is expressed as:

\[
X_i = h_i \otimes S
\]

\[
X_{ij} = \sum_{k=0}^{N-1} h_{ik} \cdot S_{j+k}
\]

where $ h_{ik} $ is the $ k $-th element of the $ i $-th column ($ i < N $) of the Hadamard matrix; $ S_{j+k} $ is the value of the time series $ S $ at position $ j+k $. $ j $ denotes the sliding window position of the convolution operation. The above formulas do not consider specific padding and dilation, which will be supplemented in the matrix form below.

The most important feature extracted from the convolution output $ X $ is the proportion of positive values (PPV):

\[
\text{ppv}_i = \frac{\sum_{j=0}^{L-N} \mathbb{I}(X_{ij} - b > 0)}{L - N + 1}
\]

where $ b $ is the bias of the convolution kernel.

\subsection{Matrix Form and Orthogonality of HIT-ROCKET}

To introduce the matrix form of HIT-ROCKET, we first need the concept of a trajectory matrix of a time series. For a time series $ S $ zero-padded to length $ l $, the trajectory matrix constructed with dimension $ d $ and delay $ \tau $ according to the Time-Delay Embedding Theorem is:

\[
X = \begin{bmatrix}
x_1 & x_{1+\tau} & \cdots & x_{1+(d-1)\tau} \\
x_2 & x_{2+\tau} & \cdots & x_{2+(d-1)\tau} \\
\vdots & \vdots & \ddots & \vdots \\
x_{m} & x_{m+\tau} & \cdots & x_{m+(d-1)\tau} \\
\end{bmatrix}
\]

where the number of rows $ m = l - (d-1)\tau $. Thus, for a convolution kernel $ h_i $ (treated as a row vector) of length $ l = N $, a convolution operation with a stride of 1 (i.e., the sliding distance of the convolution window) and dilation of 1 (dilation refers to padding zeros between kernel elements; dilation = 1 means no padding) can be expressed as:

\[
Y_i = h_i \times X
\]

Here, the number of rows of the trajectory matrix is limited to equal the length of the convolution kernel, i.e., $ m = N $. Thus, the corresponding dimension $ d = \frac{l - N}{\tau} + 1 $, where the delay $ \tau $ corresponds to the stride. If $ d $ is a fractional number, the time series $ S $ can be zero-padded to make $ d $ an integer.

For all $ N $ convolution kernels, the output is an $ N \times d $ matrix $ Y_{N \times d} $:

\[
Y_{N \times d} = H_{N \times m} X_{m \times d}
\]

Due to the orthogonality between any two column vectors $ h_i, h_j \ (i \neq j) $ of the Hadamard matrix, i.e., 

\[
\langle h_i, h_j \rangle = h_i^T h_j = \begin{cases} 
0, & i \neq j \\
\| h_i \|^2, & i = j 
\end{cases}
\]

Thus, when $ i \neq j $:

\[
(Y^T \times Y)_{ij} = \sum_{k=1}^N (X_k^T h_i^T) (h_j X_k) = \sum_{k=1}^N X_k^T (h_i^T h_j) X_k = 0
\]

When $ i = j $:

\[
(Y^T \times Y)_{ij} = \sum_{k=1}^N (X_k^T h_i^T) (h_j X_k) = \sum_{k=1}^N X_k^T (h_i^T h_j) X_k = \| h_i \|^2 = N
\]

Therefore:

\[
(Y^T \times Y)_{ii} = X^T \text{diag}(\| h_1 \|^2, \| h_2 \|^2, \ldots, \| h_d \|^2) X = N X^T X
\]

In fact, $ X^T X $ contains the system function $ A $ of the signal. For linear systems (especially when the dimension is large, the system is approximately linear locally), the system function can be estimated using the least squares method:

\[
\hat{A} = (X_1^T X_1)^{-1} X_2^T X_2
\]

where $ X_1 $ and $ X_2 $ are the first $ p-1 \ (p < d) $ columns and the last $ p-1 $ columns of the system, respectively. It can be seen that the introduction of convolution kernels does not add redundant information, thus not affecting the symplectic invariance of the system. In fact, it is equivalent to performing Walsh-Hadamard transforms on the sequence across multiple embedding dimensions (if dilated convolution is used), distributing energy evenly across all Walsh basis functions. Therefore, compared to the random convolutions in ROCKET and the non-decorrelated convolutions in miniROCKET, HIT-ROCKET achieves performance improvements while reducing computational complexity, especially for abrupt-change signals, pulse-containing signals, and piecewise constant signals.

Like all ROCKET methods, HIT-ROCKET needs to extract features such as PPV from convolution outputs, which are then input to a linear classifier for classification. For the $ i $-th convolution kernel, PPV feature extraction can be expressed as:

\[
\text{ppv}_i = \frac{\sum_{t=1}^T \mathbb{I}(y_{it} > b)}{T}
\]

where $ \mathbb{I}(x) $ is the indicator function, $ b $ is the bias, and $ T $ is the number of convolution kernel shifts.

Compared to non-decorrelated random convolution operations (e.g., Gaussian random kernels in ROCKET, 84 kernels composed of {-2, 1} in miniROCKET and MultiROCKET), the proposed Hadamard vector kernels have advantages in feature independence and robustness under noisy conditions:

\subsubsection{Feature Independence}

First, let the convolution outputs of any two different Hadamard vectors be:

\[
Y_i = H_i X, \quad Y_j = H_j X \quad (i \neq j)
\]

The covariance of the output vectors is:

\[
\text{Cov}(Y_i, Y_j) = \mathbb{E}[Y_i Y_j] = H_i^T X^T X H_j
\]

Its norm satisfies:

\[
|\text{Cov}(Y_i, Y_j)| = |H_i^T X^T X H_j| \leq |H_i| |X^T X| |H_j| \leq N |X^T X|
\]

Compared to non-orthogonal kernels in miniROCKET and MultiROCKET, whose norms are generally larger than $ N $, the covariance upper bound of Hadamard convolution outputs is smaller and only depends on the input.

For PPV features extracted from convolution outputs:

\[
\text{ppv}_i = \frac{1}{T} \sum_{t=1}^T \mathbb{I}(y_i(t) > b), \quad \text{ppv}_j = \frac{1}{T} \sum_{s=1}^T \mathbb{I}(y_j(s) > b)
\]

The covariance of the features is:

\[
\text{Cov}(\text{ppv}_i, \text{ppv}_j) = \frac{1}{T^2} \sum_{t=1}^T \sum_{s=1}^T \text{Cov}\left(\mathbb{I}(y_i(t) > b), \mathbb{I}(y_j(s) > b)\right)
\]

For binary indicator functions, the covariance is:

\[
\text{Cov}\left(\mathbb{I}(y_i(t) > b), \mathbb{I}(y_j(s) > b)\right) = P(y_i(t) > b, y_j(s) > b) - p_{i,t} p_{j,s}
\]

where $ p_{i,t} = P(y_i(t) > b) $ and $ p_{j,s} = P(y_j(s) > b) $. Assuming $ y_i(t) $ and $ y_j(s) $ follow a joint Gaussian distribution (approximated by the central limit theorem or when the input signal contains significant additive white Gaussian noise), then:

\[
P(y_i(t) > b, y_j(s) > b) = \Phi\left(\frac{b - \mu_{i,t}}{\sigma_{i,t}}, \frac{b - \mu_{j,s}}{\sigma_{j,s}}; \rho_{ij}(t,s)\right)
\]

where $ \Phi $ is the bivariate Gaussian cumulative distribution function, and $ \rho_{ij}(t,s) $ is the correlation coefficient. For zero-mean normalized outputs:

\[
\rho_{ij}(t,s) = \frac{\text{Cov}(y_i(t), y_j(s))}{N}
\]

By the Gaussian correlation inequality:

\[
P(y_i(t) > b, y_j(s) > b) \leq P(y_i(t) > b) P(y_j(s) > b) + \frac{1}{2\pi} \rho_{ij}(t,s) e^{-b^2/(2N)}
\]

Thus:

\[
\text{Cov}\left(\mathbb{I}(y_i(t) > b), \mathbb{I}(y_j(s) > b)\right) \leq \frac{1}{2\pi} |\rho_{ij}(t,s)| e^{-b^2/(2N)}
\]

The total upper bound of PPV covariance can be expressed as:

\[
\text{Cov}(\text{ppv}_i, \text{ppv}_j) \leq \frac{1}{2\pi T^2} e^{-b^2/(2N)} \sum_{t=1}^T \sum_{s=1}^T |\rho_{ij}(t,s)|
\]

where:

\[
|\rho_{ij}(t,s)| = \frac{|\text{Cov}(y_i(t), y_j(s))|}{N} = \frac{|H_i^T \Gamma(t,s) H_j|}{N}
\]

Due to the orthogonality of Hadamard vectors, off-diagonal elements of the covariance matrix $ \Gamma(t,s) = \mathbb{E}[X(t) X(s)^T] $ within the time window are suppressed, whereas non-orthogonal kernels may amplify these elements. Thus, the statistical probability of PPV features from Hadamard convolution outputs is closer to independence.

When the output PPV features are approximately statistically independent, for a total of $ M $ PPV features, if a perceptron or linear SVM is used as the classifier, the generalization error bound can be easily calculated using Hoeffding's Inequality \cite{1}:

\[
P\left(\left| \frac{1}{M} \sum_{i=1}^M \mathbb{I}(h(x_i) \neq y_i) - \frac{1}{M} \sum_{i=1}^M \mathbb{E}(\mathbb{I}(h(x_i) \neq y_i)) \right| \geq \epsilon \right) \leq 2 \exp\left(-2M\epsilon^2\right)
\]

where $ \epsilon $ is the threshold for deviation from the expected value, and $ \mathbb{I}() $ is the indicator function, denoting whether the classifier $ h(x) $ makes a prediction error on sample $ x_i $.

The sum of variances of independent random variables equals the total variance, while the sum for correlated variables is larger than the total variance. Thus, independence reduces model variance, thereby lowering the generalization error bound.

\subsubsection{Noise Robustness}

Time series signals, especially naturally occurring ones, often contain significant Gaussian noise. Let the signal-noise model be:

\[
X = X_s + X_n
\]

where $ X_n \sim \mathcal{N}(0, \sigma^2 I) $ is Gaussian noise.

The signal-to-noise ratio (SNR) of the output from convolution with Hadamard vector $ H_i $ is:

\[
\text{SNR}_{\text{Hadamard}} = \frac{\| H_i^T X_S \|^2}{\mathbb{E}[\| H_i^T N \|^2]} = \frac{(H_i^T X_S)^2}{\sigma^2 \| H_i \|^2} = \frac{(H_i^T X_S)^2}{N \sigma^2}
\]

Correspondingly, the SNR of the output from convolution with a random kernel $ K_i $ is:

\[
\text{SNR}_{\text{Correlated}} = \frac{(K_i^T S)^2}{\sigma^2 \| K_i \|^2}
\]

It can be seen that the SNR of random convolution outputs may exceed that of Hadamard convolution only if the random kernel is highly correlated with the input sequence $ X_S $ (acting as a matched filter). However, for unlearned kernels, it is difficult to achieve matched filtering for each kernel due to the inability to learn intrinsic patterns from $ X_S $. In contrast, Hadamard convolution distributes signal energy across all orthogonal channels, resulting in smaller fluctuations in thresholded features and thus stronger overall noise robustness.

HIT-ROCKET innovatively replaces the {-1, 2} combination kernels in miniROCKET with Hadamard vectors and adopts more flexible feature extraction methods, including non-uniform dilation. Below, we present the general form and feature-enhanced form of HIT-ROCKET.

\subsection{General Form}

The general form of HIT-ROCKET uses Hadamard vectors of lengths 8, 16, and 32 as convolution kernels. All elements of Hadamard convolutions are either 1 or -1, and the inner product of any two distinct Hadamard vectors is zero. Assume the length of the time series is \( N \ (N > 8) \). We select Hadamard vectors of length 8 as convolution kernels. To reduce computational complexity in the general form, no padding is applied, so the length of the convolution output sequence is \( N - 8 + 1 \). The calculation formula is as follows:

\[
y[n] = \sum_{k=0}^{7} x[n+k] h[k], \quad n = 0, 1, 2, \ldots, N-8
\]

Considering a dilation factor \( d \) for the kernel (equivalent to inserting \( d-1 \) zeros into the 8-length Hadamard vector) expands the coverage of the kernel without affecting its original orthogonality. The formula is:

\[
y[n] = \sum_{k=0}^{7} x[n+kd] h[k], \quad n = 0, 1, 2, \ldots, N-8d
\]

Compared with all other convolution methods, Hadamard convolution clearly has the lowest computational complexity. For example, using 8-length Hadamard vectors requires only 8 convolution operations with simple additions, making it highly suitable for acceleration on FPGAs \cite{22} or memristor circuits \cite{23}.

The general form of HIT-ROCKET uses only PPV (Proportion of Positive Values) as a feature, which compares convolution outputs with a series of bias values and counts how many outputs exceed these biases:

\[
\text{ppv} = \frac{\sum_{t=1}^T \mathbb{I}(y > b)}{T}
\]

where \( \mathbb{I}(x) \) is the indicator function. Here, normalization is performed using the convolution output length \( T \), but it can also be omitted.

Biases are selected randomly: during training, samples are first randomly sampled from the training set; after Hadamard convolution, quantiles of the outputs are used as biases, which are then applied to all samples.

Finally, all PPV features are directly input to various classifiers for classification. In fact, when the kernel length is \( L \in \{8, 16, 32\} \), the number of dilation factors is \( D \), the number of quantiles is \( P \), and the number of randomly sampled samples for bias generation is \( S \), the feature dimension is:

\[
f_{\text{dimension}} = L \times D \times P \times S
\]

Generally, the proposed method has far fewer features than all other ROCKET methods, including MultiROCKET, miniROCKET, and ROCKET.

\subsection{Feature-Enhanced Form}

Inspired by ROCKET and miniROCKET, more feature dimensions can be added. For example, the maximum pooling feature mentioned in ROCKET can be introduced, which takes the maximum value of Hadamard convolution outputs as a feature:

\[
\text{max} = \max(y_t)
\]

Feature extraction methods from MultiROCKET can also be adopted. First, differential sequences of the original time series can be processed simultaneously to capture higher-frequency patterns patterns using smaller-scale kernels:

\[
X' = \{ x_t - x_{t-1} : \forall t \in \{2, \ldots, N\} \}
\]

Additionally, Mean of Positive Values (MPV) can be used. MPV captures the matching strength between the input time series and a given pattern—information available but discarded when calculating PPV. It sums all positive values and computes their mean, where \( m \) is the number of positive values:

\[
\text{MPV} = \frac{1}{m} \sum_{i=1}^m y_i^+
\]

Mean of Indices of Positive Values (MIPV) may also be considered. MIPV captures information about the relative positions of positive values in convolution outputs, calculated as the mean of indices of all positive values. If an output \( Z \) has no positive values, MIPV is set to -1:

\[
\text{MIPV} = 
\begin{cases} 
\frac{1}{m} \sum_{j=1}^{m} j \mid y_j > 0 & \text{if } m > 0 \\
-1 & \text{otherwise}
\end{cases}
\]

Finally, Longest Stretch of Positive Values (LSPV) is introduced. Since MIPV aggregates all positive values, it cannot distinguish between many small sequences of consecutive positives and a few long sequences. LSPV is defined as the difference between the index of the last positive value and the index of the first positive value:

\[
\text{LSPV} = \max \{ j - i \mid \forall_{i \leq k < j}, y_k > 0 \}
\]

Examples of calculations for all four features are shown in Table \ref{tab:feature_examples}.

\begin{table}[htbp]
    \centering
    \caption{Examples of feature calculations for convolution outputs}
    \label{tab:feature_examples}
    \begin{tabular}{|c|c|c|c|c|} 
    \hline 
    Convolution outputs & PPV & MPV & MIPV & LSPV \\ 
    \hline 
    $[0, 0, 0, 0, 0, 0, 1, 1, 1, 1]$ & $0.4$ & $1$ & $7.5$ & $4$ \\ 
    \hline 
    $[1, 1, 1, 1, 0, 0, 0, 0, 0, 0]$ & $0.4$ & $1$ & $1.5$ & $4$ \\ 
    \hline 
    $[1, 1, 0, 0, 0, 0, 0, 0, 0, 1]$ & $0.4$ & $1$ & $4.5$ & $2$ \\ 
    \hline 
    $[0, 0, 0, 1, 1, 1, 1, 0, 0, 0]$ & $0.4$ & $1$ & $4.5$ & $4$ \\ 
    \hline 
    $[0, 0, 0, 0, 0, 0, 10, 10, 10, 10]$ & $0.4$ & $10$ & $7.5$ & $4$ \\ 
    \hline 
    \end{tabular}
\end{table}

Furthermore, increasing (decreasing) dilation or random dilation can be used. Increasing dilation inserts an increasing number of zeros between adjacent elements of the Hadamard vector. Specifically, for a Hadamard vector \( \mathbf{h} = [h_1, h_2, \dots, h_L] \) where \( L \in \{8, 16, 32\} \), increasing dilation inserts \( m \) zeros between the first and second elements, \( 2m \) zeros between the second and third elements, and so on. A plausible hypothesis is that such kernel transformation resembles the chirp kernel of fractional Fourier transform, leading to better performance. The vector after increasing dilation can be expressed as:

\[
\mathbf{h}_{\text{inc}} = \left[ h_1, \underbrace{0, \dots, 0}_{m \times 1}, h_2, \underbrace{0, \dots, 0}_{m \times 2}, h_3, \dots, h_{L-1}, \underbrace{0, \dots, 0}_{m \times (L-1)}, h_L \right]
\]

Random dilation inserts a random number of zeros between adjacent elements of the Hadamard vector, which may satisfy the non-uniform spacing criterion in compressed sensing, thus enhancing sparse representation capability:

\[
\mathbf{h}_{\text{rand}} = \left[ h_1, \underbrace{0, \dots, 0}_{r(1) \text{ zeros}}, h_2, \underbrace{0, \dots, 0}_{r(2) \text{ zeros}}, h_3, \dots, h_{L-1}, \underbrace{0, \dots, 0}_{r(L-1) \text{ zeros}}, h_L \right]
\]

Additionally, since Hadamard vectors have lengths that are multiples of 4, to ensure the convolution output length matches the input length after symmetric zero-padding, kernels with an odd number of elements are required. This can be achieved by inserting a zero between elements of the Hadamard vector. For example, an 8-length Hadamard vector can have a zero inserted at 9 possible positions while maintaining orthogonality.

Experiments show that the above methods can achieve higher classification accuracy on certain datasets, which will be detailed in the experimental section (Section 5).

\subsection{One-Class Classification and Anomaly Detection}

We propose introducing One-Class SVM as a classifier for anomaly detection. The goal of One-Class SVM is to find a hyperplane that separates target class data from the origin while maximizing the distance from the hyperplane to the origin. Its optimization problem is expressed as:

\begin{align*}
\min_{\mathbf{w}, b, \xi} \quad & \frac{1}{2} \|\mathbf{w}\|^2 + \frac{1}{\nu N} \sum_{i=1}^N \xi_i \\
\text{subject to} \quad & \mathbf{w}^T \psi(\mathbf{x}_i) + b \geq 1 - \xi_i, \quad \xi_i \geq 0, \quad i = 1, 2, \ldots, N
\end{align*}

where:
\begin{itemize}
    \item $\mathbf{w}$: Weight vector.
    \item $\psi(\mathbf{x}_i)$: Kernel function mapping input data to a high-dimensional space.
    \item $b$: Bias term.
    \item $\xi_i$: Slack variables for handling noise and outliers.
    \item $\nu$: Regularization parameter controlling the proportion of outliers.
\end{itemize}

Furthermore, we use LS-OCSVM and RLS-OCSVM proposed in \cite{24}, which adopt the squared loss $(\mathbf{w}^T \psi(\mathbf{x}_i) + b)^2$ to handle noisy data. They also incorporate penalty terms for the mean and variance of modeling errors in the optimization problem to enhance robustness. The optimization problem is:

\[
\min_{\mathbf{w}, b} \quad \frac{1}{2} \|\mathbf{w}\|^2 + \frac{\gamma}{2} \sum_{i=1}^{N} (\mathbf{w}^T \psi(\mathbf{x}_i) + b)^2 + \frac{\lambda}{2} \left( \frac{1}{N} \sum_{i=1}^{N} (\mathbf{w}^T \psi(\mathbf{x}_i) + b) \right)^2
\]

where:
\begin{itemize}
    \item $\mathbf{w}$: Weight vector.
    \item $\psi(\mathbf{x}_i)$: Kernel function mapping input data to a high-dimensional space.
    \item $b$: Bias term.
    \item $\gamma$: Regularization parameter controlling the weight of squared loss.
    \item $\lambda$: Regularization parameter controlling the penalty term for the mean of modeling errors.
\end{itemize}

\subsection{Pseudocode of the Main HIT-ROCKET Version}
\begin{algorithm}[H]
\caption{Hadamard Convolution Transform Rocket (HIT-ROCKET)}
\begin{algorithmic}[1]
\REQUIRE 
\STATE \quad \( \mathcal{D}_{\text{train}} = \{ (\mathbf{x}_i, y_i) \}_{i=1}^N \) (training time series dataset), 
\STATE \quad \( \mathcal{D}_{\text{test}} = \{ \mathbf{x}_j \}_{j=1}^M \) (test time series dataset),
\STATE \quad \( L \in \{8, 16, 32\} \) (kernel lengths), 
\STATE \quad \( D = \{d_1, d_2, \ldots, d_k\} \) (dilation factors),
\STATE \quad \( P \) (number of quantiles for biases), 
\STATE \quad \( S \) (number of random samples for bias generation),
\STATE \quad \( C \) (classifier, e.g., logistic regression)

\ENSURE 
\STATE \quad \( \hat{y}_j \in \mathcal{Y} \) (predictions for test dataset)

\STATE \textbf{Step 1: Generate Hadamard kernels}
\FOR{each \( l \in L \)}
    \STATE Generate \( l \)-length Hadamard matrix \( H_l \in \{\pm1\}^{l \times l} \)
    \STATE Extract columns of \( H_l \) as kernels: \( \mathbf{h}_1, \mathbf{h}_2, \ldots, \mathbf{h}_l \in \{\pm1\}^l \)
\ENDFOR

\STATE \textbf{Step 2: Generate bias values}
\STATE Randomly sample \( S \) time series from \( \mathcal{D}_{\text{train}} \)
\STATE Initialize temporary set \( \mathcal{Y} \leftarrow \emptyset \)
\FOR{each sampled time series \( \mathbf{x} \)}
    \FOR{each kernel \( \mathbf{h} \in \{\mathbf{h}_1, \ldots, \mathbf{h}_l\} \)}
        \FOR{each dilation \( d \in D \)}
            \STATE Compute dilated convolution: \( y = \mathbf{h} \otimes_d \mathbf{x} \)
            \STATE Add \( y \) to \( \mathcal{Y} \)
        \ENDFOR
    \ENDFOR
\ENDFOR
\STATE Compute \( P \) quantiles of \( \mathcal{Y} \) as biases: \( b_1, b_2, \ldots, b_P \)

\STATE \textbf{Step 3: Extract training features}
\FOR{each \( (\mathbf{x}_i, y_i) \in \mathcal{D}_{\text{train}} \)}
    \STATE Initialize feature vector \( \mathbf{f}_i \leftarrow [] \)
    \FOR{each kernel \( \mathbf{h} \)}
        \FOR{each dilation \( d \in D \)}
            \STATE Compute \( y = \mathbf{h} \otimes_d \mathbf{x}_i \)
            \FOR{each bias \( b_p \in \{b_1, \ldots, b_P\} \)}
                \STATE Calculate PPV: \( \text{ppv} = \frac{1}{|y|} \sum_{t=1}^{|y|} \mathbb{I}(y[t] > b_p) \)
                \STATE Append \( \text{ppv} \) to \( \mathbf{f}_i \)
            \ENDFOR
        \ENDFOR
    \ENDFOR
\ENDFOR

\STATE \textbf{Step 4: Train classifier}
\STATE Train \( C \) on \( \{ (\mathbf{f}_i, y_i) \}_{i=1}^N \)

\STATE \textbf{Step 5: Predict on test data}
\FOR{each \( \mathbf{x}_j \in \mathcal{D}_{\text{test}} \)}
    \STATE Extract feature vector \( \mathbf{f}_j \) using Step 3
    \STATE Predict \( \hat{y}_j = C(\mathbf{f}_j) \)
\ENDFOR

\RETURN \( \{ \hat{y}_j \}_{j=1}^M \)
\end{algorithmic}
\end{algorithm}

\section{EXPERIMENTS}
\clearpage % 强制换页
\subsection{Experimental Configuration}

All numerical experiments were programmed using Python and conducted on a system with a 13th Gen Intel(R) Core(TM) i5 13600KF CPU and an RTX 4080S GPU. GPU acceleration via the PyTorch library was used during the Hadamard convolution transformation process. A CPU computation version based on the NumPy library is also provided, though there may be room for further optimization.

The performance of the model was first validated using the well-known UCR dataset. The UCR (2019) dataset contains 128 subfolders, each representing a distinct time series classification task from various domains, with pre-defined training and test set splits. Overall, the number of samples in the training set is smaller than that in the test set. We evaluated both the general form and feature-enhanced form of HIT-ROCKET, comparing it with key competitors: miniROCKET and MultiROCKET. During comparisons, we ensured the number of features was as close as possible. For comparisons with miniROCKET, features were generated at three levels: 0.5k, 3k, and 8k. Relevant parameters are shown in Tables 2 and 3.

\begin{table}[htbp]
    \centering
    \caption{Parameters for HIT-ROCKET feature generation}
    \begin{tabular}{|c|c|c|c|c|}
    \hline
    Feature Level / Actual Count & Dilation List & Bias Quantiles & Sampled Samples & Kernel Length \\
    \hline
    0.5k / 538 & range(1, 12) & [0.618] & 3 & 16 \\
    \hline
    3k / 3200 & range(1, 21) & [0.618, 0.95] & 5 & 16 \\
    \hline
    8k / 7680 & range(1, 25) & [0.05, 0.3, 0.618, 0.95] & 5 & 16 \\
    \hline
    \end{tabular}
\end{table}

\begin{table}[htbp]
    \centering
    \caption{Parameters for miniROCKET feature generation}
    \begin{tabular}{|c|c|c|c|}
    \hline
    Feature Level / Actual Count & Dilation List & Bias Quantiles & Sampled Samples \\
    \hline
    0.5k / 504 & range(1, 7) & [0.95] & 1 \\
    \hline
    3k / 3780 & range(1, 10) & [0.95] & 5 \\
    \hline
    8k / 7560 & range(1, 10) & [0.05, 0.3, 0.618, 0.95] & 5 \\
    \hline
    \end{tabular}
\end{table}

Additionally, additive white Gaussian noise was added to the original UCR dataset (including test and validation sets) such that the signal-to-noise ratio (SNR) varied from 20 dB to 50 dB in 5 dB increments. Under these conditions, HIT-ROCKET and miniROCKET were trained, and the mean and variance of their F1 scores were compared to evaluate noise robustness.

The UCR dataset was also used for time series one-class detection tasks. Features of each class were extracted using miniROCKET, MultiROCKET, and HIT-ROCKET. The class with the largest number of samples was designated as the positive class, and all other classes as negative classes. One-Class SVM, RLS-OCSVM, and SRLS-OCSVM from scikit-learn were used for one-class classification in the backend.

Finally, non-uniform dilation and differencing methods were used to expand feature dimensions (while only using PPV features), and performance was compared with MultiROCKET.

\subsection{Time Series Classification Experiments: Comparison with miniROCKET}

\begin{figure}[htbp]
    \centering
    \includegraphics[width=1\linewidth]{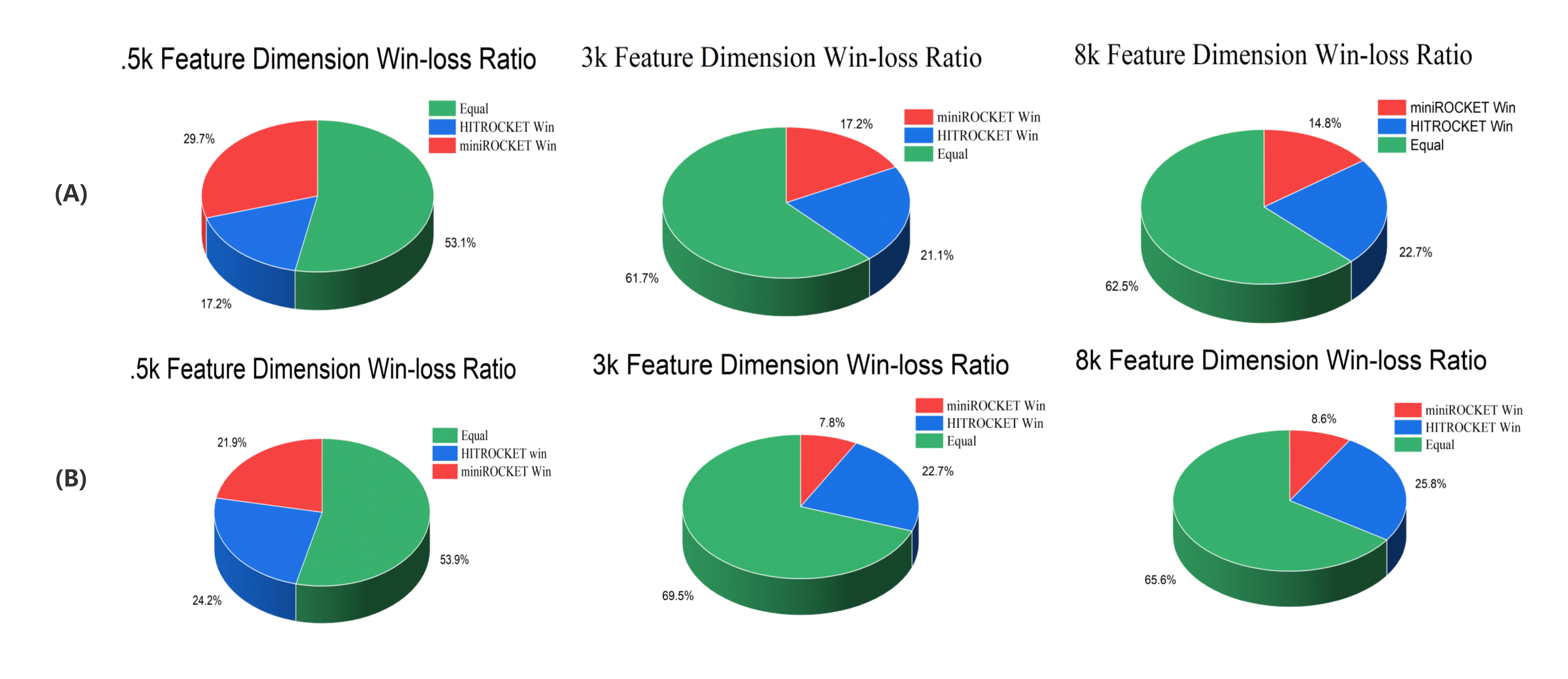}
    \caption{\small (A) shows the classification performance comparison between HIT-ROCKET and miniROCKET using a ridge regression classifier at the three feature levels (0.5K, 3K, 8K); (B) displays the comparison using a logistic regression classifier at the same feature levels; (C) and (D) present the variance of F1 scores for HIT-ROCKET and miniROCKET under ridge regression and logistic regression, respectively, as SNR varies from 20 dB to 50 dB.}
    \label{fig:2}
\end{figure}

We evaluated time series classification tasks across 128 subfolders in the UCR dataset. Using the original dataset splits (including training/test sample counts and sequence lengths), HIT-ROCKET achieved better performance with less training time in most cases. Since bias generation requires random sampling from samples, each dataset was trained 10 times, and the final F1 score was taken as the mean of the results. For the test sets of the 128 subfolders, comparisons between HIT-ROCKET and its main competitor miniROCKET were conducted at three feature levels (0.5K, 3K, and 8K) using ridge regression and logistic regression classifiers. The results are shown in \ref{fig:2}.

As observed from the figures, when the feature dimension exceeds 1K, HIT-ROCKET's classification accuracy is higher than miniROCKET's. In addition to the orthogonality mentioned in \textit{Preliminaries}, potential reasons include: with fixed output feature dimensions, HIT-ROCKET uses fewer kernels, allowing more hyperparameter combinations to enhance feature extraction capability. Furthermore, using logistic regression as the classifier yields more significant performance improvements than ridge regression.

Overall, for the 128 UCR datasets, the following table presents the proportion of F1 scores in different ranges:

\begin{table}[htbp]
    \centering
    \caption{Distribution of F1 scores across different classifiers and feature levels}
    \begin{tabular}{|c|c|c|c|c|c|}
    \hline
    Classifier & \multicolumn{3}{c|}{.5K} & 3K & 8K \\
    \hline
    & F1$\geq$0.85 & 0.85$>$F1$>$0.65 & F1$<$0.65 & F1$\geq$0.85 & 0.85$>$F1$>$0.65 \\
    \hline
    Ridge Regression & 51 & 45 & 32 & 66 & 42 \\
    \hline
    Logistic Regression & 59 & 42 & 27 & 69 & 44 \\
    \hline
    SVM & 60 & 42 & 25 & 68 & 42 \\
    \hline
    Tree Models & 24 & 44 & 56 & 24 & 44 \\
    \hline
    \end{tabular}
\end{table}

As shown in the table, linear models achieve performance close to SVM-based models, while tree models perform poorly. Considering training time and computational costs, ridge regression and logistic regression are recommended. The SVM category includes linear kernel SVM, RBF kernel SVM, sigmoid kernel SVM, and polynomial kernel SVM, with the best model selected during training. Experiments show that linear and RBF kernels achieve good performance on most datasets. For tree models, the best model was selected from decision trees, boosted trees, and random forests. Random forests performed best in most cases, but due to high training time and deployment costs, results were only obtained for 0.5k and 3k feature dimensions.

\begin{figure}[H]
    \centering
    \includegraphics[width=1\linewidth]{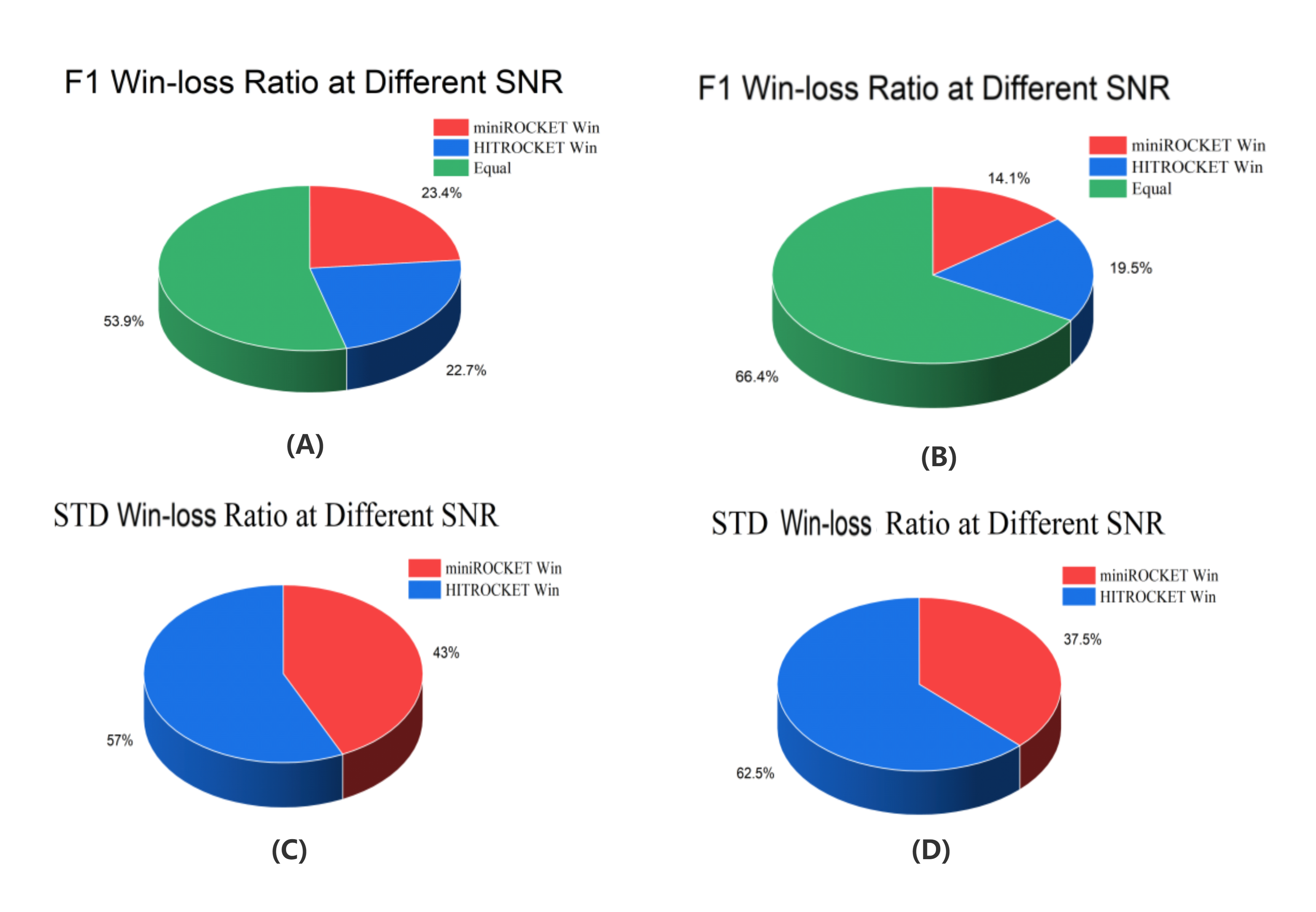} % 替换为实际图片文件名
    \caption{\small (A) Average F1 score comparison using ridge regression; (B) Average F1 score comparison using logistic regression; (C) Variance of F1 scores for ridge regression; (D) Variance of F1 scores for logistic regression. HIT-ROCKET consistently shows more stable performance across different SNR levels.}
    \label{fig:3}
\end{figure}

To further verify that HIT-ROCKET has better noise robustness than the baseline model (miniROCKET), we added noise to the training and test sets of the 128 folders with SNR values of 20, 25, 30, ..., 50 dB. Using 3k feature dimensions, comparisons between the proposed method and miniROCKET showed that: in most cases, HIT-ROCKET and miniROCKET achieved similar average F1 scores with ridge regression, while HIT-ROCKET achieved higher average scores with logistic regression. However, HIT-ROCKET exhibited smaller variance in scores as SNR changed. The detailed results are presented in Figure \ref{fig:3}.

\subsection{Training Time}

\begin{figure}[H]
    \centering
    \includegraphics[width=1\linewidth]{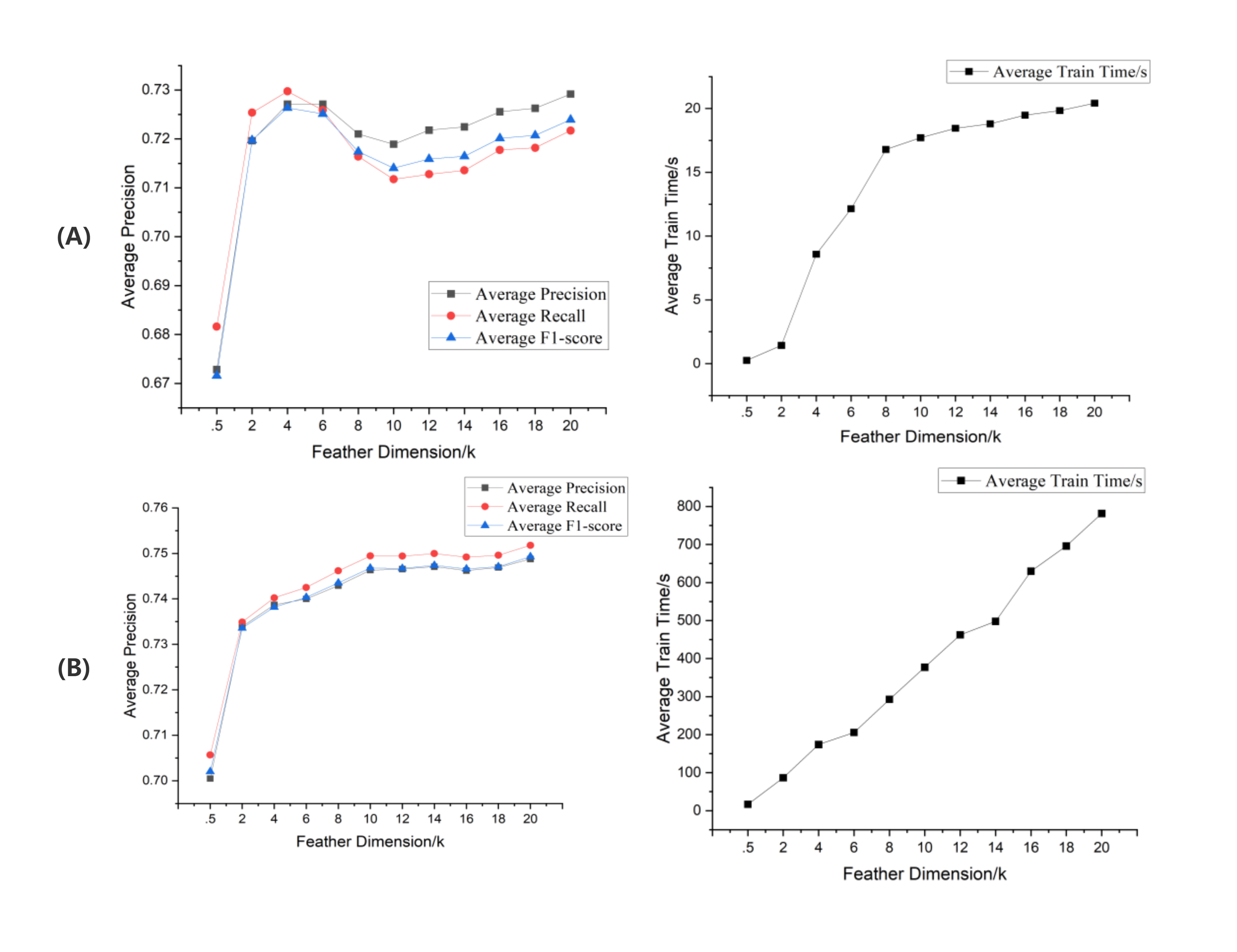} 
    \caption{\small (A) shows the average training F1 scores (left) and average training times (right) of HIT-ROCKET as functions of feature dimension using a ridge regression classifier; (B) displays the same metrics using a logistic regression classifier.}
    \label{fig:4}
\end{figure}

With fixed feature dimensions, training time is slightly faster than miniROCKET. However, with fixed hyperparameter combinations, computational costs are significantly reduced due to fewer kernels. PyTorch-accelerated conversion code is provided. Experiments were conducted on the UCR dataset with the largest number of samples, using logistic regression and ridge regression classifiers. Feature dimensions varied from 0.5k to 20k, with each dimension trained 5 times to calculate the average F1 score. The model's training time and F1 score as functions of feature dimension are shown in \ref{fig:4}.

\subsection{One-Class Classification}

Features were extracted from the UCR dataset using miniROCKET and the proposed HIT-ROCKET method. For each dataset, the class with the most samples was designated as the normal class. One-class classification was performed using One-Class SVM (linear and RBF kernels) and RLS-OCSVM from the scikit-learn library, with experiments conducted at 0.5k and 3k feature dimensions. The performance results are shown in the following table:

\begin{table}[htbp]
    \centering
    \caption{One-class classification performance across different classifiers and feature dimensions}
    \begin{tabular}{|c|c|c|c|c|}
    \hline
    Classifier & \multicolumn{3}{c|}{.5K Feature Dimensions} & 3K Feature Dimensions \\
    \hline
    & F1$\geq$0.85 & 0.85$>$F1$>$0.65 & F1$<$0.65 & F1$\geq$0.85 \\
    \hline
    Linear-OCSVM (Sklearn) & 14 & 16 & 94 & 18 \\
    \hline
    RBF-OCSVM (Sklearn) & 13 & 16 & 95 & 18 \\
    \hline
    RLS-OCSVM & 53 & 38 & 33 & 51 \\
    \hline
    \end{tabular}
\end{table}

Notably, when using RLS-OCSVM as the classifier for one-class tasks, the recall metric was excellent, demonstrating the robustness and efficiency of the proposed algorithm. Combining HIT-ROCKET with RLS-OCSVM (known for robustness) achieved superior results, as shown in the following table (datasets with insufficient samples for training were excluded):

\begin{table}[htbp]
    \centering
    \caption{Recall performance of one-class classification across different classifiers and feature dimensions}
    \begin{tabular}{|c|c|c|c|c|}
    \hline
    Classifier & \multicolumn{3}{c|}{.5K Feature Dimensions} & 3K Feature Dimensions \\
    \hline
    & R$\geq$0.85 & 0.85$>$R$>$0.65 & R$<$0.65 & R$\geq$0.85 \\
    \hline
    Linear-OCSVM (Sklearn) & 13 & 10 & 101 & 14 \\
    \hline
    RBF-OCSVM (Sklearn) & 9 & 13 & 102 & 15 \\
    \hline
    RLS-OCSVM & 123 & 1 & 0 & 123 \\
    \hline
    \end{tabular}
\end{table}

\subsection{Feature Enhancement}

Inspired by the chirp transform basis of fractional Fourier transform, we extended HIT-ROCKET's feature dimensions using increasing dilation convolutions, differential sequence processing, and only PPV features. At the same feature dimensions (5k, 10k, 20k), performance was compared with MultiROCKET using a ridge regression classifier on 8 randomly selected datasets from the UCR dataset. The full results are shown in the following table:

% 第一部分表格：包含5K和10K特征维度
\begin{table}[htbp]
    \centering
    \footnotesize
    \setlength{\tabcolsep}{4pt}
    \caption{Performance comparison between MultiROCKET and enhanced HITrocket (mul-HITrocket) across different feature dimensions (Part 1)}
    \begin{tabular}{|l|c|c|c|c|}
    \hline
    \multirow{2}{*}{Dataset} & \multicolumn{2}{c|}{5K Feature Dimensions} & \multicolumn{2}{c|}{10K Feature Dimensions} \\
    \cline{2-5}
    & MultiROCKET & mul-HITrocket & MultiROCKET & mul-HITrocket \\
    \hline
    CricketY & 0.77 & 0.80 & 0.82 & 0.79 \\
    \hline
    CricketZ & 0.76 & 0.77 & 0.81 & 0.79 \\
    \hline
    EOGHorizontalSignal & 0.42 & 0.47 & 0.43 & 0.52 \\
    \hline
    FiftyWords & 0.70 & 0.79 & 0.74 & 0.80 \\
    \hline
    FordB & 0.77 & 0.81 & 0.78 & 0.80 \\
    \hline
    GestureMidAirD1 & 0.71 & 0.72 & 0.76 & 0.72 \\
    \hline
    OliveOil & 0.93 & 0.89 & 0.89 & 0.89 \\
    \hline
    Worms & 0.72 & 0.80 & 0.77 & 0.76 \\
    \hline
    \end{tabular}
    \label{tab:feature_comparison_part1}
\end{table}

% 第二部分表格：包含20K特征维度和基线
\begin{table}[htbp]
    \centering
    \footnotesize
    \setlength{\tabcolsep}{4pt}
    \caption{Performance comparison between MultiROCKET and enhanced HITrocket (mul-HITrocket) across different feature dimensions (Part 2)}
    \begin{tabular}{|l|c|c|c|}
    \hline
    \multirow{2}{*}{Dataset} & \multicolumn{2}{c|}{20K Feature Dimensions} & \multirow{2}{*}{\tiny Baseline (8K)} \\ % 表头缩小
    \cline{2-3}
    & MultiROCKET & mul-HITrocket & \tiny MiniROCKET \\ % 对应内容也缩小
    \hline
    CricketY & 0.83 & 0.79 & 0.77 \\
    \hline
    CricketZ & 0.80 & 0.79 & 0.83 \\
    \hline
    EOGHorizontalSignal & 0.44 & 0.52 & 0.38 \\
    \hline
    FiftyWords & 0.74 & 0.78 & 0.70 \\
    \hline
    FordB & 0.80 & 0.79 & 0.77 \\
    \hline
    GestureMidAirD1 & 0.72 & 0.71 & 0.68 \\
    \hline
    OliveOil & 0.93 & 0.93 & 0.93 \\
    \hline
    Worms & 0.78 & 0.81 & 0.76 \\
    \hline
    \end{tabular}
    \label{tab:feature_comparison_part2}
\end{table}

The results show that feature-enhanced HIT-ROCKET achieves better classification performance than the non-enhanced version and meets or exceeds MultiROCKET (the enhanced version of miniROCKET). Each experiment was conducted once, so scores may contain some randomness.
\section{DISCUSSION}

First, some suggestions for hyperparameter tuning are provided. HITrocket requires tuning hyperparameters including the length of Hadamard convolution kernels, bias quantiles, number of dilations, number of random samples, and classifier type. Through extensive tuning experiments, we found that a Hadamard convolution vector length of 16 achieves a good balance between performance and computational efficiency. For bias quantiles, we recommend selecting 0.95 or 0.05 to indicate strong or weak correlation between the kernel and input sequences, respectively. The number of dilations should range from 1 to approximately 20, and the number of random samples within 20. Regarding the tuning order, we suggest adjusting the number of random samples first, followed by quantiles and dilations. Alternatively, grid search methods can be used. For the final classification model, considering the balance between performance and efficiency, we recommend prioritizing ridge regression, logistic regression, and support vector machines with linear or RBF kernels. For one-class classification tasks, RLS-SVM is a better choice.

Second, in terms of computational complexity, since the convolution kernels only contain 1 and -1, related operations can be completely converted to addition operations. Due to the small number of kernels, more hyperparameter combinations can be accommodated, thereby achieving better classification performance than methods like MiniROCKET with fewer feature dimensions. Furthermore, the main work has accelerated the feature transformation process using CUDA via PyTorch. Codes using NumPy and C++ with Eigen are also provided, though the optimization of the latter may not be optimal.

There is room for further improvement of the proposed method. Future work will focus on:

1. Currently, we have shown that our method outperforms state-of-the-art methods such as GRU, TCN, and Transformer on certain datasets in further experiments, with more experiments being supplemented.
2. Research on multi-input scenarios: we are currently investigating whether to process each input independently or treat them as multivariate data. Future work can further explore \textbf{mixed processing strategies}, dynamically selecting processing methods based on input correlation and importance, while combining hierarchical processing and dynamic input selection mechanisms to better mine relationships between inputs and improve model flexibility.
3. Design of more efficient algorithm frameworks: subsequent research will explore using lightweight classifiers such as binary neural networks to replace the current statistical learning-based classifiers, aiming to further improve computational efficiency.
4. Parallel computing kernels based on the OpenCL 4.x specification are under development to support acceleration on embedded platform devices such as NPU, DSP, and FPGA.

\section{CONCLUSIONS}

This paper proposes the HITrocket time series classification algorithm for time series classification tasks, which currently achieves the highest computational efficiency while reaching state-of-the-art classification accuracy, making it suitable for deployment on edge devices with minimal computing power such as microcontrollers. The proposed HITrocket method uses Hadamard vectors as convolution kernels, reducing computational complexity while ensuring the independence of feature extraction. As a result, it achieves better classification performance than its main competitors (MiniROCKET and MultiROCKET) under the same hyperparameter combinations and feature dimensions, with stronger noise robustness. Additionally, we further explored one-class time series classification tasks and classification tasks with a large number of classes. The superiority of our method across various tasks has been demonstrated on both classic UCR datasets and chaos-generated datasets.

\bibliographystyle{unsrt} % 数字编号的参考文献样式
\bibliography{sample} % 关联你的.bib文件
\end{document}